\documentclass[10pt,twocolumn,letterpaper]{article}

\usepackage{wacv}

\usepackage{graphicx}
\usepackage{amsmath}
\usepackage{amssymb}
\usepackage{booktabs}

\usepackage[pagebackref,breaklinks,colorlinks]{hyperref}

\usepackage[capitalize]{cleveref}
\crefname{section}{Sec.}{Secs.}
\Crefname{section}{Section}{Sections}
\Crefname{table}{Table}{Tables}
\crefname{table}{Tab.}{Tabs.}

\def\confName{WACV}
\def\confYear{2024}

\begin{document}

\title{Super Efficient Neural Network for Compression Artifacts Reduction and Super Resolution}

\author{Wen Ma, Qiuwen Lou, Arman Kazemi, Julian Faraone, Tariq Afzal\\
Amazon Lab126\\
1100 Enterprise Way, Sunnyvale, CA 94089\\
{\tt\small wwenma@amazon.com}\\
}
\maketitle

\begin{abstract}
   Video quality can suffer from limited internet speed while being streamed by users. Compression artifacts start to appear when the bitrate decreases to match the available bandwidth. Existing algorithms either focus on removing the compression artifacts at the same video resolution, or on upscaling the video resolution but not removing the artifacts. Super resolution-only approaches will amplify the artifacts along with the details by default. We propose a lightweight convolutional neural network (CNN)-based algorithm which simultaneously performs artifacts reduction and super resolution (ARSR) by enhancing the feature extraction layers and designing a custom training dataset. The output of this neural network is evaluated for test streams compressed at low bitrates using variable bitrate (VBR) encoding. The output video quality shows a 4-6 increase in video multi-method assessment fusion (VMAF) score compared to traditional interpolation upscaling approaches such as Lanczos or Bicubic.
\end{abstract}

\section{Introduction}
\label{sec:intro}

Video streaming has become a major entertainment activity in many households around the world. With the recent popularity of AI/ML hardware accelerators, neural network-based upscaler has been of interests to be integrated on edge devices like 4K displays, laptops, mobile phones, TVs, to achieve real time upscaling of video contents. One of the challenges is that when the internet bandwidth/bitrate is low, the input videos suffer from compression artifacts such as blocking, ringing, flickering artifacts, color changes, etc. Machine learning and neural network-based algorithms have been widely studied to address the problem of removing compression artifacts in videos, mainly based on convolutional neural networks (CNNs)~\cite{Authors21}~\cite{Authors22}. Usually these models are heavy weight and not feasible to run on resource constrained edge and mobile devices, as they use multiple frames as the input to produce one output frame, and often involves complicated algorithms such as optic flow and motion compensation. In this paper, we introduce a hardware friendly network that simultaneously performs artifacts reduction and super resolution (ARSR), by combining super efficient neural network (SESR)~\cite{Authors10}, a lightweight CNN-based solution to super resolution, with an artifacts reduction approach used in ARCNN~\cite{Authors12}. The network can produce output video quality close to a state-of-the-art model, BasicVSR++, for video super resolution (VSR) and video enhancement (VE) at a significantly reduced model size. 

The major contributions of this paper are: 1) proposing a lightweight CNN-based model with 22K parameters for ARSR. 2) using only one frame at a time to produce one output upscaled/enhanced frame, 3) using over-parameterization on both feature extraction and non-linear mapping layers to improve the inference efficiency and picture quality, and 4) using video multi-method assessment fusion (VMAF) to evaluate video quality in test datasets.

\section{Background}

During the past few years, the trend for image restoration and enhancement has been pivoted from single image super resolution~\cite{Authors01} to video super resolution and real world video problems, especially on video enhancement and quality improvement on compressed videos. BasicVSR++~\cite{Authors02}, a winner of the NTIRE challenge held in 2021, has been treated as a state-of-the-art model for video super resolution and compressed video enhancement. The subsequent winners of this challenge~\cite{Authors03} are heavily adapted from this model. BasicVSR++ is based on recurrent network and uses optical flow that uses second order grid propagation and flow guided deformable alignment, both of which involve complicated algorithms. In addition, the model has 7.3M parameters and cannot be easily implemented in edge devices.

Another popular architecture to address the video super resolution and video enhancement problems is the generative adversarial networks (GAN)~\cite{Authors04}. The advantage of using GAN is the ability to use little input data to generate realistic outputs. A GAN architecture is made of two networks, a discriminator network that distinguishes ground truth high resolution images from upscaled ones, and a generator network that produces images that gradually look like ground truth images. There is a notable work called SUPERVEGAN~\cite{Authors05} that uses GAN network for video super resolution. SUPERVEGAN combines dynamic upsampling filters (DUF) and GAN to perform artifacts removal and super resolution at the same time. However, it is using 5 low resolution input frames to generate a single high resolution frame. In addition, the model is developed initially as a software approach to address the VSR and VE problems and is not suitable for hardware implementations.

Super Resolution algorithms based on simple stacking of convolution layers is favored due to the potential simplicity of the algorithms and models. Several lightweight models have been proposed, such as FSRCNN~\cite{Authors06}, ABRL~\cite{Authors07}, LapSRN~\cite{Authors08}, CARN-M~\cite{Authors09}, etc. The number of weight parameters can be reduced to less than 30K in some of these models and the amount of MACs can be reduced to less 100 billion when upscaling images from 360p to 720p resolution. A notable solution in this category is the super efficient super solution (SESR) network~\cite{Authors10}, which uses over-parameterization and collapsible linear blocks to reduce the inference complexity, making it suitable for hardware inference operations. However, SESR only performs super resolution but not artifacts reduction. Recently, SRCNN~\cite{Authors11} model architecture has been modified to ARCNN~\cite{Authors12} to deal with compression artifacts removal. The number of convolutional layers in ARCNN was expanded for feature extraction and feature enhancement. The non-linear mapping and reconstruction layers remained the same. Inspired by~\cite{Authors12}, we came up with the idea of extending SESR for artifacts reduction by adopting similar techniques used for ARCNN. Details will be elaborated in section 3.

Video multi-method assessment fusion, or VMAF~\cite{Authors13} was developed for perceptual video quality assessment. The other video quality metrics such as PSNR or SSIM that use pixel-based calculations may fail to capture human perception accurately~\cite{Authors14}. VMAF, on the other hand, combines human vision modeling with support vector machines to produce a score that represents human perception. The final metric score is calculated using the weighted scores from several elementary metrics. The elementary metrics scores were obtained using opinion scores through subjective experiments. We adopted this method as it is open sourced and can be directly used along with FFmpeg commands. For measuring video quality shown on 4K TVs, we used the VMAF model for 4K resolution while disabling enhancement gain mode (NEG).

\begin{figure}[t]
  \centering
   \includegraphics[width=0.8\linewidth]{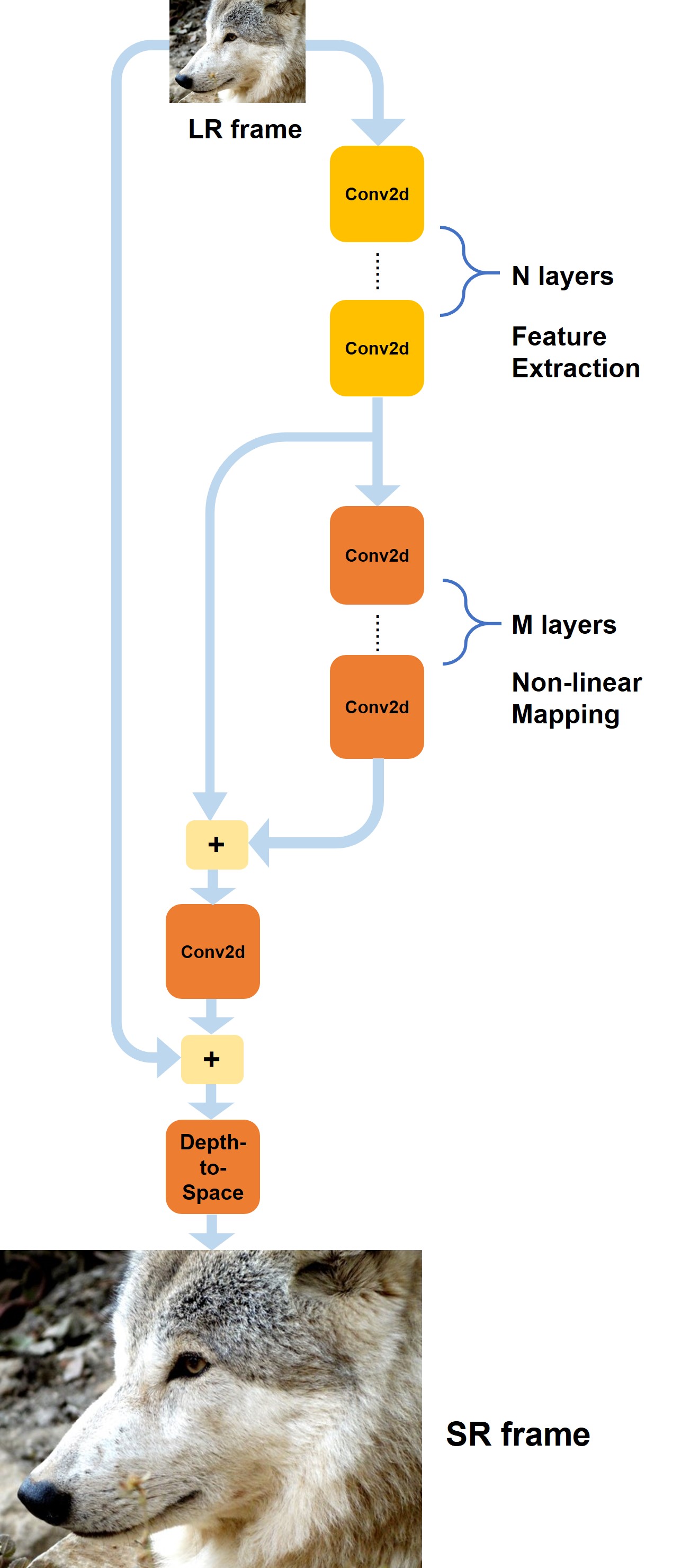}
   \caption{Model architecture of the ARSR network. The low resolution (LR) frame goes through $N$ Conv2d layers for feature extraction, $M$ Conv2d layers for non-linear mapping, a Conv2d layer to match the upscaling factor, and a final depth-to-space layer to upscale the frame to super resolution (SR). For later experiments, unless specified, $N = 3$ and $M = 11$ are used.}
   \label{fig:1}
\end{figure}

\section{Super Efficient Neural Network for Artifacts Reduction and Super Resolution}
\label{sec:formatting}

Our neural network architecture was adapted from SESR~\cite{Authors10} which does not perform artifacts reduction and focuses on single image super resolution. Various modifications of the neural network have been tested to optimize the performance, such as architecture change, training dataset modifications, loss function variations, different chroma channel upscaling techniques. Also the network was studied to support different upscaling factors such as $\times 2$, $\times 3$ and $\times 4$. To make the network hardware friendly, techniques such as grouped convolutions, quantization aware training, power-of-two scaling were tested on video enhancement tasks.

\subsection{Overall ARSR Neural Network Architecture}

The architecture of the CNN-based ARSR Network is shown in \cref{fig:1}. The network only processes the Y channel component of single images. There is a global residual layer feeding the input image (Y channel only) to the depth-to-space layer at the end. On the other side, the input image goes through $N$ layers of feature extraction. A second residual layer starts from the beginning of non-linear mapping layers and ends at the end of the non-linear mapping layers. The number of non-linear mapping layers ($M$ in \cref{fig:1}) can range from 5 to 11 depending on the image quality or hardware cost requirements. The depth-to-space layer at the end performs the actual upscaling of images, therefore relaxing the memory requirements because the intermediate outputs of previous layers are at the low resolution (LR) image sizes.

\begin{figure}[t]
  \centering
   \includegraphics[width=0.98\linewidth]{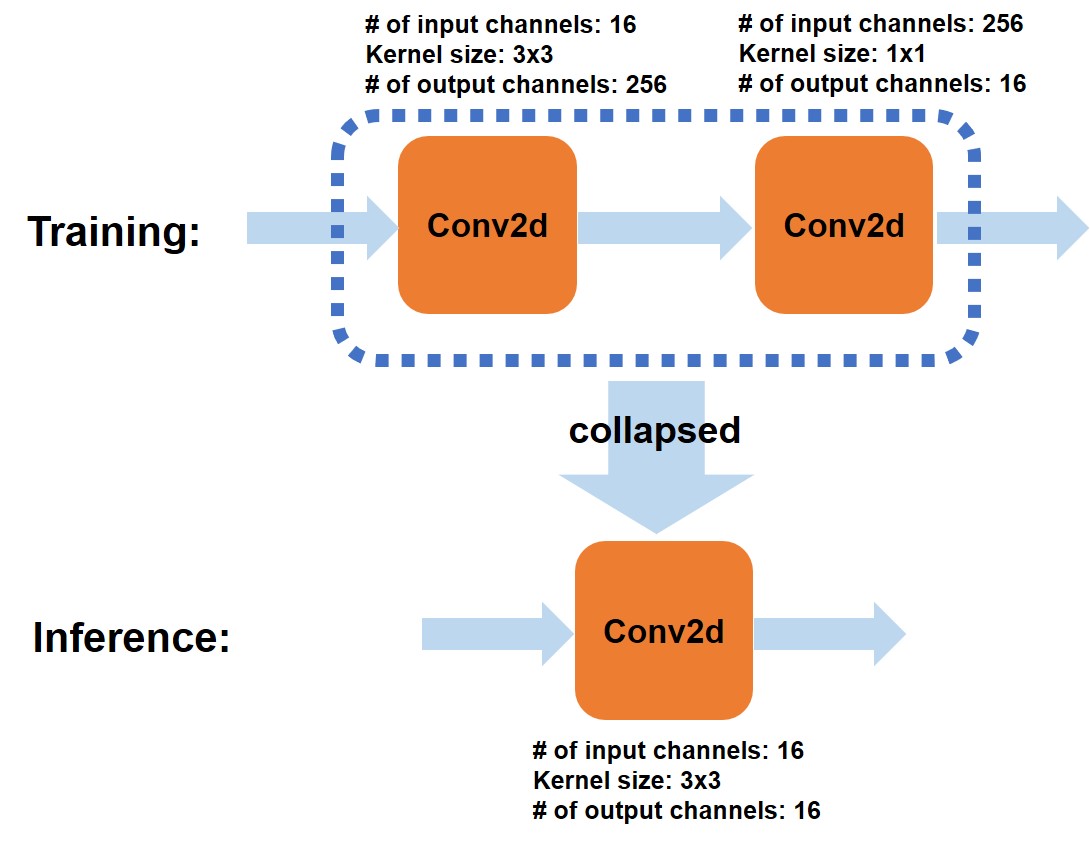}
   \caption{During training, the Conv2d operation is expanded into two sequential Conv2d operations. This is an example where the number of input and output channels are both 16 while the number of internal expanded channel is 256.}
   \label{fig:2}
\end{figure}

The Conv2d layers all use over-parameterization techniques where a normal convolution operation with $n$ input channels, $m$ output channels, and $f \times f$ kernel sizes will be expanded to two convolution operations where the first one has $n$ input channels, $p$ output channels, with kernel size of $f \times f$, while the second one has $p$ input channels, $m$ output channels, with kernel  size of $1 \times 1$. Usually the value of $p$ is much larger than $m$ and $n$. During training, the weight updates are done on the expanded network, with $npff+pm$ weights. During the inference, only the $nmff$ collapsed weights are used, leading to more much efficient inference operations. The over-parameterization scheme is shown in \cref{fig:2}.

\subsection{Training Dataset Generation}

\begin{figure}
  \centering
  \begin{subfigure}{0.9\linewidth}
    \includegraphics[width=0.95\linewidth]{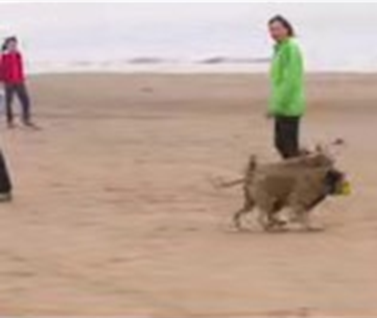}
    \caption{Low resolution (LR) image for training, compressed at 50 kbps using H.265 codec.}
    \label{fig:3-1}
  \end{subfigure}
  \vfill
  \begin{subfigure}{0.9\linewidth}
    \includegraphics[width=0.95\linewidth]{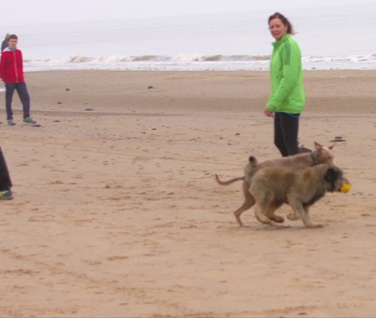}
    \caption{High resolution (HR) image for training.}
    \label{fig:3-2}
  \end{subfigure}
  \caption{Examples from the Vimeo dataset for training.}
  \label{fig:3}
\end{figure}

In stead of using the traditional bicubic downscaled images as the low resolution samples in the training dataset, we compressed and downscaled the original videos from the Vimeo dataset~\cite{Authors15} using H.265 codec at a low bitrate - 50 kbps. We used extracted frames of the 4K resolution videos from the Vimeo dataset as the high resolution samples, and the extracted frames of the compressed and downscaled videos as the low resolution samples. Examples of the low resolution and high resolution video frames are shown in \cref{fig:3}. We also tried another method of injecting JPEG compression artifacts into the traditional bicubic downscaled low resolution samples but this method tends to smooth out everything in the output image. Our adopted method can smooth out area where there are artifacts and still keep the high frequency details.

\begin{figure*}
  \centering
  \begin{subfigure}{0.23\linewidth}
    \includegraphics[width=0.99\linewidth]{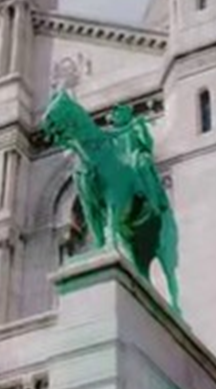}
    \caption{Input 540p frame with compression artifacts.}
    \label{fig:4-1}
  \end{subfigure}
  \hspace{0.3em}
  \begin{subfigure}{0.23\linewidth}
    \includegraphics[width=0.99\linewidth]{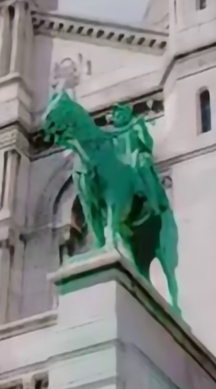}
    \caption{Output 4K frame trained with MAE loss.}
    \label{fig:4-2}
  \end{subfigure}
  \hspace{0.3em}
  \begin{subfigure}{0.23\linewidth}
    \includegraphics[width=0.99\linewidth]{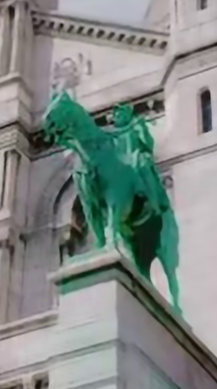}
    \caption{ Output 4K frame trained with MSE loss.}
    \label{fig:4-3}
  \end{subfigure}
  \hspace{0.3em}
  \begin{subfigure}{0.23\linewidth}
    \includegraphics[width=0.99\linewidth]{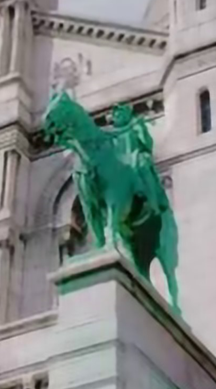}
    \caption{Output 4K frame trained with Huber loss.}
    \label{fig:4-4}
  \end{subfigure}
  \caption{Comparison between different loss functions during training.}
  \label{fig:4}
\end{figure*}

\begin{figure*}
  \centering
  \begin{subfigure}{0.23\linewidth}
    \includegraphics[width=0.99\linewidth]{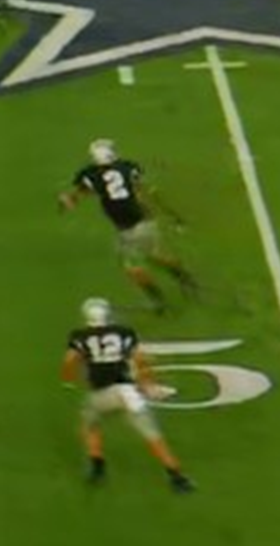}
    \caption{Input 540p frame with compression artifacts.}
    \label{fig:5-1}
  \end{subfigure}
  \hspace{0.3em}
  \begin{subfigure}{0.23\linewidth}
    \includegraphics[width=0.99\linewidth]{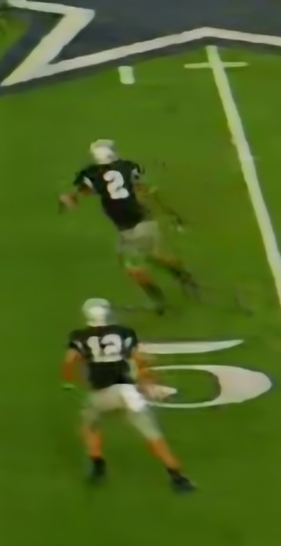}
    \caption{Upscaled 4K frame with 1 layer of feature extraction.}
    \label{fig:5-2}
  \end{subfigure}
  \hspace{0.3em}
  \begin{subfigure}{0.23\linewidth}
    \includegraphics[width=0.99\linewidth]{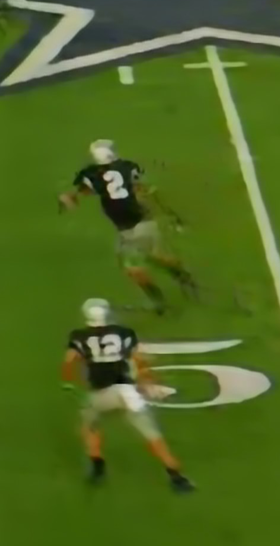}
    \caption{Upscaled 4K frame with 2 layers of feature extraction.}
    \label{fig:5-3}
  \end{subfigure}
  \hspace{0.3em}
  \begin{subfigure}{0.23\linewidth}
    \includegraphics[width=0.99\linewidth]{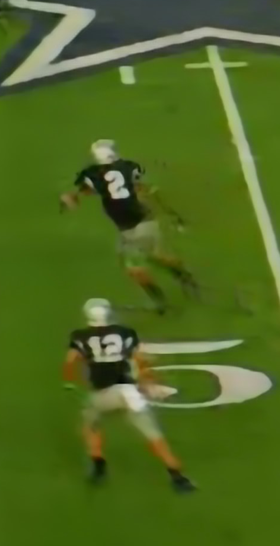}
    \caption{Upscaled 4K frame with 3 layers of feature extraction.}
    \label{fig:5-4}
  \end{subfigure}
  \caption{Using different number of layers of feature extraction to reduce artifacts.}
  \label{fig:5}
\end{figure*}

The dataset generated is based on single frame, meaning that one input image frame is used to generate one output frame. It is not uncommon that multiple input frames are used in CNN to generate one output frame ~\cite{Authors16} for video super resolution tasks. We explored the multi-frame approach by using three input frames as the input and the high resolution version of the middle frame as the output. The result looks almost the same as the single frame approach.

\begin{figure*}
  \centering
  \begin{subfigure}{0.31\linewidth}
    \includegraphics[width=0.99\linewidth]{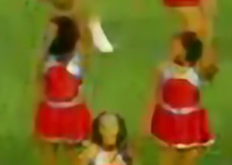}
    \caption{Using Bilinear interpolation for chroma channels.}
    \label{fig:6-1}
  \end{subfigure}
  \hspace{0.3em}
  \begin{subfigure}{0.31\linewidth}
    \includegraphics[width=0.99\linewidth]{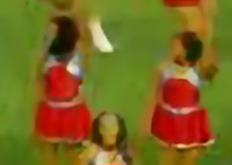}
    \caption{Using Bicubic interpolation for chroma channels.}
    \label{fig:6-2}
  \end{subfigure}
  \hspace{0.3em}
  \begin{subfigure}{0.31\linewidth}
    \includegraphics[width=0.99\linewidth]{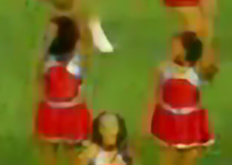}
    \caption{Using Nearest Neighbor interpolation for chroma channels.}
    \label{fig:6-3}
  \end{subfigure}
  \caption{Using different chroma channels upscaling methods.}
  \label{fig:6}
\end{figure*}

\begin{table*}
  \centering
  \begin{tabular}{p{0.15\textwidth}p{0.15\textwidth}p{0.15\textwidth}p{0.15\textwidth}p{0.15\textwidth}}
    \toprule
    Input Resolution & Output Resolution & Upscale Factor & Number of Output Channels in the Last Conv2d Layer & Use of Lanczos upscaler or not\\
    \midrule
    $960 \times 540$ & $3840 \times 2160$ & 4 & 16 & No \\
    $1280 \times 720$ & $3840 \times 2160$ & 3 & 9 & No \\
    $1920 \times 1080$ & $3840 \times 2160$ & 2 & 4 & No \\
    $960 \times 540$ & $2560 \times 1440$ & 8/3 & 4 & Yes \\
    $1280 \times 720$ & $2560 \times 1440$ & 2 & 4 & No \\
    $2560 \times 1440$ & $3840 \times 2160$ & 3/2 & N/A & Yes \\
    \bottomrule
  \end{tabular}
  \caption{ARSR network can support different upscaling factors combining lanczos upscaler when needed.}
  \label{tab:1}
\end{table*}

\subsection{Loss Function Exploration}

Different kinds of loss functions during training have been explored, including the mean average error (MAE), mean squared error (MSE), and Huber loss. We found that MAE performs the best at artifacts reduction while Huber loss and MSE preserves slightly more details at the cost of not removing enough artifacts. A comparison between using different loss functions during training is shown in \cref{fig:4}.

\subsection{Expanding Feature Extraction Layers for Artifacts Reduction}

Inspired by~\cite{Authors12}, we increased the number of feature extraction layers ($N$ in \cref{fig:1}) in the beginning of the network from one layer to multiple layers to perform better artifacts reduction. More feature extraction layers can extract more features from the input or training images to construct better output images and produce better training results. For three layers of feature extraction, we adopted the $7 \times 7$ kernel size in the first layer, followed by $5 \times 5$ and $3 \times 3$ kernel size in the second and third layers. For two layers of feature extraction, we adopted the $7 \times 7$ kernel size in the first layer, followed by $5 \times 5$ kernel size in the second layer. The number of input and output channels are both 16. We found that using two or three layers of feature extraction is sufficient to remove the artifacts, as shown in \cref{fig:5}.

\subsection{Chorma Channels Upscaling Methods}

While the ARSR neural network is used to upscale Luma channel (Y channel) only, traditional upscaling method is used to upscale the chroma channels (Cb, Cr channels). From~\cite{Authors11}, we know that using neural network-based upscaler on luma channel only is sufficient to generate competitive super resolution results. We explored using bicubic, bilinear and nearest neighbor methods for chroma upscaling, and found that the bilinear method produces the smoothest edges, while nearest neighbor method tend to add some artifacts near the edges (\cref{fig:6}). The nearest neighbor method is the easiest to implement in hardware as it has the lowest computation complexity, whereas the bicubic method is the most difficult. We have decided to use bilinear upscaling for chroma channels because it can remove the jaggedness on edges and produce smooth-looking outputs.

\begin{figure*}
  \centering
  \begin{subfigure}{0.23\linewidth}
    \includegraphics[width=0.99\linewidth]{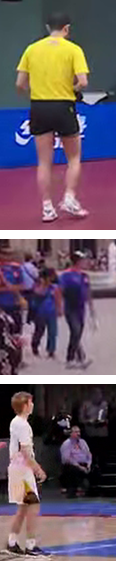}
    \caption{Upscaled 4K frame by Lanczos interpolation.}
    \label{fig:7-1}
  \end{subfigure}
  \hspace{0.1em}
  \begin{subfigure}{0.23\linewidth}
    \includegraphics[width=0.99\linewidth]{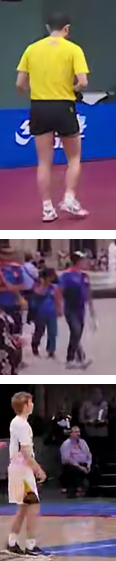}
    \caption{Upscaled 4K frame by ARSR quantized network.}
    \label{fig:7-2}
  \end{subfigure}
  \hspace{0.1em}
  \begin{subfigure}{0.23\linewidth}
    \includegraphics[width=0.99\linewidth]{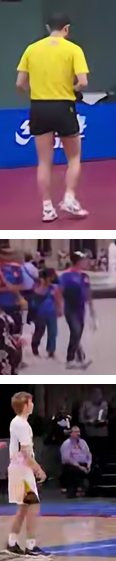}
    \caption{Upscaled 4K frame by ARSR FP32 network.}
    \label{fig:7-3}
  \end{subfigure}
  \hspace{0.1em}
  \begin{subfigure}{0.23\linewidth}
    \includegraphics[width=0.99\linewidth]{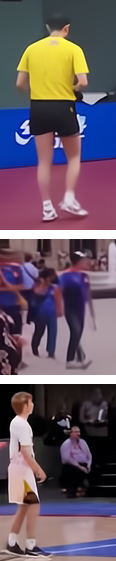}
    \caption{Upscaled 4K frame by BasicVSR++.}
    \label{fig:7-4}
  \end{subfigure}
  \caption{Comparison between different upscaling methods.}
  \label{fig:7}
\end{figure*}

\begin{figure*}
  \centering
  \begin{subfigure}{0.23\linewidth}
    \includegraphics[width=0.99\linewidth]{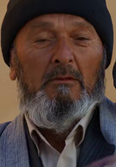}
    \caption{Upscaled 4K frame by Lanczos interpolation.}
    \label{fig:8-1}
  \end{subfigure}
  \hspace{0.1em}
  \begin{subfigure}{0.23\linewidth}
    \includegraphics[width=0.99\linewidth]{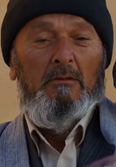}
    \caption{Upscaled 4K frame by ARSR quantized network.}
    \label{fig:8-2}
  \end{subfigure}
  \hspace{0.1em}
  \begin{subfigure}{0.23\linewidth}
    \includegraphics[width=0.99\linewidth]{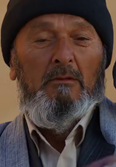}
    \caption{Upscaled 4K frame by ARSR FP32 network.}
    \label{fig:8-3}
  \end{subfigure}
  \hspace{0.1em}
  \begin{subfigure}{0.23\linewidth}
    \includegraphics[width=0.99\linewidth]{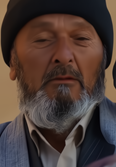}
    \caption{Upscaled 4K frame by BasicVSR++.}
    \label{fig:8-4}
  \end{subfigure}
  \caption{One example showing too much artifacts reduction can lead to loss of details.}
  \label{fig:8}
\end{figure*}

\begin{table*}
  \centering
  \begin{tabular}{p{0.3\textwidth}p{0.3\textwidth}p{0.3\textwidth}}
    \toprule
    Upscaling Method & Average VMAF of 8 test videos & Number of Parameters\\
    \midrule
    ARSR FP32 & 52.89 & 41.2K \\
    ARSR quantized & 51.64 & 22.2K\\
    Lanczos & 47.55 & N/A\\
    Bicubic & 45.37 & N/A\\
    BasicVSR++ & 55.68 & 7.3M \\
    \bottomrule
  \end{tabular}
  \caption{VMAF score comparison when upscaling 540p compressed videos to 4K videos. ARSR network is producing higher VMAF scores than Lanzcos and Bicubic upscalers.}
  \label{tab:2}
\end{table*}

We have also tried to modify the neural network to be 3-channel models, either based on YCbCr or RGB. We found that 3-channel models can produce slightly better output image quality but there is color difference between the expected output and generated output. Therefore we didn't pursue this method further.

\subsection{Supporting Different Upscaling Scenarios}

The ARSR neural network can support $\times2$, $\times3$, and $\times4$ upscaling factors by changing the number of output channels in the last Conv2d layer shown in \cref{fig:1}, before the depth-to-space layer. If the upscaling factor is $n$, the number of output channels would be $n^2$. So a larger integer upscaling factor would require slightly more parameters in the neural network. \cref{tab:1} shows different upscaling scenarios for different input video resolutions and output video resolutions. For upscaling scenarios where the upscaling factor is not an integer, we upscale using the biggest integer that will result in an output resolution lower than the desired output resolution and then use the lanczos interpolation method to upscale to the desired output resolution.

\subsection{Techniques to Improve the Efficiency of ARSR Neural Network}

To reduce the number of parameters and computations in the network, we propose to use the grouped convolutions~\cite{Authors18} to divide the 16 input and 16 output channels of the non-linear mapping layers into $n$ groups. The value of $n$ can be 8, 4 or 2. Normal Conv2d operations are performed within each group of input channels and output channels. The outputs from each group are finally concatenated into the final output. The number of model parameters and computations of the non-linear mapping layers is reduced by $n$.

Other techniques we adopted to make the network suitable for hardware implementation include quantization-aware training~\cite{Authors19} and power-of-two scaling factor during quantization~\cite{Authors20}. Power-of-two scaling during quantization can convert the multiplication operation during quantization/dequantization into bit-shift operations to simplify the computation. However, using power-of-two scale factors can lead to lower accuracy due to the rounding errors so we didn't integrate this technique into our quantized model.

\section{Experiment Results of the ARSR Neural Network}

Instead of using the popular PSNR or SSIM metrics to evaluate picture quality, we use the VMAF score to measure quality as it can better evaluate the degree of artifacts reduction in addition to detail enhancement. The model of VMAF\_4K\_NEG is used because it measures the audience's response when they are viewing the video on a 4K TV. We selected 8 videos from the inter4K dataset~\cite{Authors17} and generated a testing dataset for ARSR evaluation. We compressed the video using variable bitrate (VBR) encoding at 500 kbps and then upscale by ARSR network or other comparable methods. The baseline we chose are lanczos, which is the most complicated traditional interpolation method, and bicubic upscaling approaches.

We demonstrated the results of ARSR network by showing the output of a quantized $\times4$ network that uses group of 4 convolutions and 12 bit on weights and activations. The test inputs are 540p compressed videos. The test outputs are upscaled and enhanced 4K videos. In comparison, the outputs of the ARSR FP32 (Floating Point 32-bit) network, the state-of-the-art BasicVSR++ network, and baseline Lanczos upscaler are shown as well (\cref{fig:7}). We see that the Lanczos upscaler does not remove compression artifacts or mosquito noise, while the BasicVSR++ performs the best at removing artifacts. However, BasicVSR++ can sometimes remove too much artifacts that it smooths out the details, as shown in \cref{fig:8}. The outputs of the quantized ARSR network is slightly worse than the FP32 network, but still better than the Lanczos upscaled outputs (\cref{fig:7}). 

To evaluate the video quality quantitatively, we measured the VMAF scores of the 8 upscaled output videos and calculated the average. The result is shown in \cref{tab:2}. BasicVSR++ has the highest VMAF score, which indicates the best video quality, followed by ARSR FP32 network and ARSR quantized network. On the other hand, BasicVSR++ also uses the most complex model with 7.3M parameters, whereas ARSR FP32 has 41.2K parameters and ARSR quantized only has 22.2K parameters. The Lanczos and Bicubic methods have the lowest VMAF scores. We believe this table has shown that our ARSR network and training method is effective in producing noticeable improvement in video quality.

\section{Conclusion}

We have developed a CNN-based neural network for simultaneous artifacts removal and super resolution (ARSR). It is lightweight, based on single frame, and easy to implement in hardware. Based on a selected test dataset, the performance of ARSR is about 4-6 higher in the VMAF score than lanczos or bicubic methods. The state-of-the-art model, BasicVSR++ which has 7.3M parameters, is not feasible to run on resource constrained edge device hardware. This is the first known hardware friendly network that performs artifacts reduction and super resolution at the same time.

Some of the future work can include recovering the picture quality loss from the quantized model, making different training dataset for different input video bitrate/resolution scenarios, and further reducing the network size by decreasing the number of non-linear mapping layers, e.g. $M=5$.


\begin{thebibliography}{10}\itemsep=-1pt

\bibitem{Authors09}
Namhyuk Ahn, Byungkon Kang, and Kyung-Ah Sohn.
\newblock Fast, accurate, and lightweight super-resolution with cascading residual network.
\newblock In {\em Computer Vision – ECCV 2018: 15th European Conference, Munich, Germany, September 8-14, 2018, Proceedings, Part X}, page 256–272, Berlin, Heidelberg, 2018. Springer-Verlag.

\bibitem{Authors05}
Silviu~S. Andrei, Nataliya Shapovalova, and Walterio Mayol-Cuevas.
\newblock Supervegan: Super resolution video enhancement gan for perceptually improving low bitrate streams.
\newblock In {\em IEEE Access}, volume~9, pages 91160--91174, 2021.

\bibitem{Authors10}
Kartikeya Bhardwaj, Milos Milosavljevic, Liam O\textquotesingle~Neil, Dibakar Gope, Ramon Matas, Alex Chalfin, Naveen Suda, Lingchuan Meng, and Danny Loh.
\newblock Collapsible linear blocks for super-efficient super resolution.
\newblock In D. Marculescu, Y. Chi, and C. Wu, editors, {\em Proceedings of Machine Learning and Systems}, volume~4, pages 529--547, 2022.

\bibitem{Authors02}
Kelvin~C.K. Chan, Shangchen Zhou, Xiangyu Xu, and Chen~Change Loy.
\newblock Basicvsr++: Improving video super-resolution with enhanced propagation and alignment.
\newblock In {\em 2022 IEEE/CVF Conference on Computer Vision and Pattern Recognition (CVPR)}, pages 5962--5971, 2022.

\bibitem{Authors22}
Wei-Gang Chen, Runyi Yu, and Xun Wang.
\newblock Neural network-based video compression artifact reduction using temporal correlation and sparsity prior predictions.
\newblock {\em IEEE Access}, 8:162479--162490, 2020.

\bibitem{Authors12}
Chao Dong, Yubin Deng, Chen~Change Loy, and Xiaoou Tang.
\newblock Compression artifacts reduction by a deep convolutional network.
\newblock In {\em 2015 IEEE International Conference on Computer Vision (ICCV)}, pages 576--584, 2015.

\bibitem{Authors11}
Chao Dong, Chen~Change Loy, Kaiming He, and Xiaoou Tang.
\newblock Image super-resolution using deep convolutional networks.
\newblock {\em IEEE Trans. Pattern Anal. Mach. Intell.}, 38(2):295–307, feb 2016.

\bibitem{Authors06}
Chao Dong, Chen~Change Loy, and Xiaoou Tang.
\newblock Accelerating the super-resolution convolutional neural network.
\newblock In Bastian Leibe, Jiri Matas, Nicu Sebe, and Max Welling, editors, {\em Computer Vision -- ECCV 2016}, pages 391--407, Cham, 2016. Springer International Publishing.

\bibitem{Authors07}
Z. Du, J. Liu, J. Tang, and G. Wu.
\newblock Anchor-based plain net for mobile image super-resolution.
\newblock In {\em 2021 IEEE/CVF Conference on Computer Vision and Pattern Recognition Workshops (CVPRW)}, pages 2494--2502, Los Alamitos, CA, USA, jun 2021. IEEE Computer Society.

\bibitem{Authors04}
Ian Goodfellow, Jean Pouget-Abadie, Mehdi Mirza, Bing Xu, David Warde-Farley, Sherjil Ozair, Aaron Courville, and Yoshua Bengio.
\newblock Generative adversarial nets.
\newblock In {\em Advances in neural information processing systems}, pages 2672--2680, 2014.

\bibitem{Authors19}
Benoit Jacob, Skirmantas Kligys, Bo Chen, Menglong Zhu, Matthew Tang, Andrew Howard, Hartwig Adam, and Dmitry Kalenichenko.
\newblock Quantization and training of neural networks for efficient integer-arithmetic-only inference.
\newblock In {\em Proceedings of the IEEE conference on computer vision and pattern recognition}, pages 2704--2713, 2018.

\bibitem{Authors16}
Armin Kappeler, Seunghwan Yoo, Qiqin Dai, and Aggelos~K Katsaggelos.
\newblock Video super-resolution with convolutional neural networks.
\newblock {\em IEEE transactions on computational imaging}, 2(2):109--122, 2016.

\bibitem{Authors18}
Alex Krizhevsky, Ilya Sutskever, and Geoffrey~E Hinton.
\newblock Imagenet classification with deep convolutional neural networks.
\newblock In F. Pereira, C.J. Burges, L. Bottou, and K.Q. Weinberger, editors, {\em Advances in Neural Information Processing Systems}, volume~25. Curran Associates, Inc., 2012.

\bibitem{Authors08}
Wei-Sheng Lai, Jia-Bin Huang, Narendra Ahuja, and Ming-Hsuan Yang.
\newblock Deep laplacian pyramid networks for fast and accurate super-resolution.
\newblock In {\em 2017 IEEE Conference on Computer Vision and Pattern Recognition (CVPR)}, pages 5835--5843, 2017.

\bibitem{Authors13}
Zhi Li, Anne Aaron, Ioannis Katsavounidis, Anush Moorthy, Megha Manohara, et~al.
\newblock Toward a practical perceptual video quality metric.
\newblock {\em The Netflix Tech Blog}, 6(2):2, 2016.

\bibitem{Authors14}
Juan~Carlos Mier, Eddie Huang, Hossein Talebi, Feng Yang, and Peyman Milanfar.
\newblock Deep perceptual image quality assessment for compression.
\newblock In {\em 2021 IEEE International Conference on Image Processing (ICIP)}, pages 1484--1488. IEEE, 2021.

\bibitem{Authors20}
Dominika Przewlocka-Rus, Syed~Shakib Sarwar, H~Ekin Sumbul, Yuecheng Li, and Barbara De~Salvo.
\newblock Power-of-two quantization for low bitwidth and hardware compliant neural networks.
\newblock {\em arXiv preprint arXiv:2203.05025}, 2022.

\bibitem{Authors21}
Jae~Woong Soh, Jaewoo Park, Yoonsik Kim, Byeongyong Ahn, Hyun-Seung Lee, Young-Su Moon, and Nam~Ik Cho.
\newblock Reduction of video compression artifacts based on deep temporal networks.
\newblock {\em IEEE Access}, 6:63094--63106, 2018.

\bibitem{Authors17}
Alexandros Stergiou and Ronald Poppe.
\newblock Adapool: Exponential adaptive pooling for information-retaining downsampling.
\newblock {\em IEEE Transactions on Image Processing}, 32:251--266, 2022.

\bibitem{Authors01}
Radu Timofte, Shuhang Gu, Jiqing Wu, and Luc Van~Gool.
\newblock Ntire 2018 challenge on single image super-resolution: Methods and results.
\newblock In {\em Proceedings of the IEEE conference on computer vision and pattern recognition workshops}, pages 852--863, 2018.

\bibitem{Authors15}
Tianfan Xue, Baian Chen, Jiajun Wu, Donglai Wei, and William~T Freeman.
\newblock Video enhancement with task-oriented flow.
\newblock {\em International Journal of Computer Vision}, 127:1106--1125, 2019.

\bibitem{Authors03}
Ren Yang, Radu Timofte, Meisong Zheng, Qunliang Xing, Minglang Qiao, Mai Xu, Lai Jiang, Huaida Liu, Ying Chen, Youcheng Ben, et~al.
\newblock Ntire 2022 challenge on super-resolution and quality enhancement of compressed video: Dataset, methods and results.
\newblock In {\em Proceedings of the IEEE/CVF Conference on Computer Vision and Pattern Recognition}, pages 1221--1238, 2022.

\end{thebibliography}
\end{document}